\documentclass[letterpaper]{article} 
\usepackage{aaai2026}  
\usepackage{times}  
\usepackage{helvet}  
\usepackage{courier}  
\usepackage[hyphens]{url}  
\usepackage{graphicx} 
\usepackage{natbib}  
\usepackage{caption} 
\usepackage{algorithm}
\usepackage{algorithmic}
\usepackage{amsmath} 
\usepackage{booktabs}
\usepackage{multirow}
\usepackage{amssymb}

\usepackage{newfloat}
\usepackage{listings}
\DeclareCaptionStyle{ruled}{labelfont=normalfont,labelsep=colon,strut=off} 
\floatstyle{ruled}
\newfloat{listing}{tb}{lst}{}
\floatname{listing}{Listing}
\title{ICM-Fusion: In-Context Meta-Optimized LoRA Fusion for Multi-Task Adaptation}
\author{
    Yihua Shao\textsuperscript{\rm 1,2}, 
    Xiaofeng Lin\textsuperscript{\rm 3}, 
    Xinwei Long\textsuperscript{\rm 4}, 
    Siyu Chen\textsuperscript{\rm 2}, 
    Minxi Yan\textsuperscript{\rm 5}, 
    Yang Liu\textsuperscript{\rm 6}, \\
    Ziyang Yan\textsuperscript{\rm 7}, 
    Ao Ma\textsuperscript{\rm 8}, 
    Hao Tang\textsuperscript{\rm 9}, 
    Jingcai Guo\textsuperscript{\rm 1}\thanks{Corresponding author: jc-jingcai.guo@polyu.edu.hk}
}
\affiliations{
    \textsuperscript{\rm 1}The Hong Kong Polytechnic University,
    \textsuperscript{\rm 2}Institute of Automation, Chinese Academy of Sciences,\\
    \textsuperscript{\rm 3}	Guangdong University of Technology,
    \textsuperscript{\rm 4}Tsinghua University,
    \textsuperscript{\rm 5}The Chinese University of Hong Kong,\\
    \textsuperscript{\rm 6}	Beijing Institute for General Artificial Intelligence,
    \textsuperscript{\rm 7}	University of Trento,\\
    \textsuperscript{\rm 8}	JD,
    \textsuperscript{\rm 9}	Peking University

}

\usepackage{bibentry}

\begin{document}

\maketitle

\begin{abstract}

Enabling multi-task adaptation in pre-trained Low-Rank Adaptation (LoRA) models is crucial for enhancing their generalization capabilities. Most existing pre-trained LoRA fusion methods decompose weight matrices, sharing similar parameters while merging divergent ones. However, this paradigm inevitably induces inter-weight conflicts and leads to catastrophic domain forgetting. While incremental learning enables adaptation to multiple tasks, it struggles to achieve generalization in few-shot scenarios. Consequently, when the weight data follows a long-tailed distribution, it can lead to forgetting in the fused weights. To address this issue, we propose \textbf{In-Context Meta LoRA Fusion (ICM-Fusion)}, a novel framework that synergizes meta-learning with in-context adaptation. The key innovation lies in our task vector arithmetic, which dynamically balances conflicting optimization directions across domains through learned manifold projections. ICM-Fusion obtains the optimal task vector orientation for the fused model in the latent space by adjusting the orientation of the task vectors. Subsequently, the fused LoRA is reconstructed by a self-designed \textbf{Fusion VAE (F-VAE)} to realize multi-task LoRA generation. We have conducted extensive experiments on visual and linguistic tasks, and the experimental results demonstrate that ICM-Fusion can be adapted to a wide range of architectural models and applied to various tasks. Compared to the current pre-trained LoRA fusion method, ICM-Fusion fused LoRA can significantly reduce the multi-tasking loss and can even achieve task enhancement in few-shot scenarios.

\end{abstract}

\section{Introduction}
Multiple LoRA fusion enables efficient multi-task adaptation for large language models after training \cite{shao2024gwq, shao2025eventvad}. Existing methods~\cite{zhang2023adalora,wang2024lora} predominantly rely on SVD~\cite{koren2008factorization} low-rank decomposition for information interpolation. However, these approaches cause damage to LoRA's intrinsic structure during interpolation. Shared-weight methods can mitigate this degradation~\cite{huang2023compeft}, but incur knowledge forgetting when handling significantly divergent weights~\cite{zhang2024dlp}. While VAE~\cite{kingma2013auto} sampling offers information protection, and methods like ICM-LoRA~\cite{shao2025context} have attempted model fusion in latent states, they still incur substantial information loss at deeper levels. Therefore, we introduce meta-learning~\cite{hospedales2021meta} into the inter-weight space, enhancing individual LoRA weights' capacity for multi-task adaptation.

\begin{figure}
    \centering
    \includegraphics[width=0.95\linewidth]{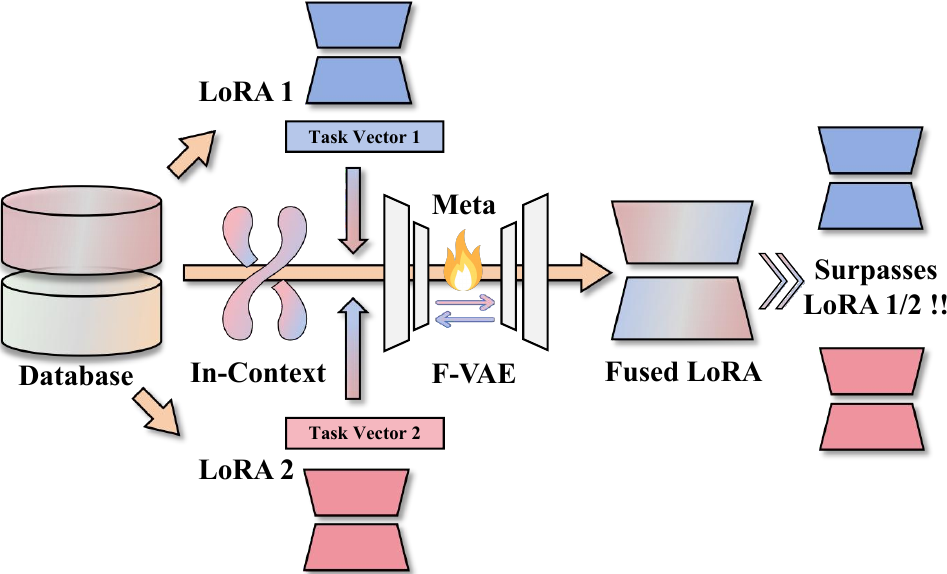}
    \caption{ICM-Fusion enables multi-task generation of LoRA parameters via dynamic task vector arithmetic and context modeling, enhancing model generalization across diverse domains and few-shot scenarios.}
    \label{fig:enter-label}
\end{figure}

Meanwhile, incremental learning~\cite{gepperth2016incremental,yan2024renderworld,masana2022class} and MoE (Mixture of Experts)~\cite{liao2025gm,zoph2022st} reduce knowledge forgetting in multi-task LoRA training. Yet their generalizability in few-shot scenarios remains limited~\cite{jiang2024few, yan2023nerfbk}. For instance, MoE-LoRA~\cite{wu2024mixture} significantly mitigates forgetting but fails to address data scarcity in few-shot tasks. Furthermore, incremental learning and MoE approaches only achieve adaptation within their predefined task scope, struggling to enhance model capability on out-of-domain generalization tasks~\cite{shao2024accidentblip,wu2024meta}.

To address this, we propose the \textbf{In-Context Meta LoRA Fusion} method, which introduces a Fusion Variational Autoencoder (VAE) framework combined with contextual meta-learning to achieve seamless fusion of LoRA parameters. Task vectors, extracted from the hidden states of pre-trained models, guide the VAE in encoding parameters into a unified latent space, aligning distributions to reduce conflicts and forgetting. Meanwhile, contextual meta-learning optimizes the latent representations, enhancing the model's generalization ability and enabling rapid adaptation to new tasks with low computational overhead.

Our main contributions can be summarized as follows:  
\begin{itemize}
    \item We propose a novel framework that efficiently merges multiple LoRA models without relying on original datasets, significantly improving parameter storage and inference efficiency.
    \item Through semantic guidance from task vectors, we achieve alignment and fusion of task-specific parameters, alleviating model conflicts and online forgetting.
    \item Our method demonstrates superior performance over baseline approaches in multi-task scenarios while maintaining low computational complexity and resource demands.
\end{itemize}

\section{Related Work}
\subsection{Model Fusion}
Model fusion has emerged as a critical area of research in machine learning, enabling the efficient combination of multiple models into a single, more powerful and efficient model~\cite{li2025gvd,li2023deep,tang2024fusionbench, wang2025unifying, shao2025tr}. Early works, such as Model Soups~\cite{wortsman2022model}, demonstrated that averaging weights of independently trained models can improve out-of-distribution (OOD) performance~\cite{jang2024model,wortsman2022robust, guo2025gp}. Other foundational techniques include Singular Value Decomposition (SVD)~\cite{koren2008factorization} for model compression and fusion, with extensions like SVD in Latent State that optimize merging in reduced-dimensional spaces. Regularization-based methods such as RegMean~\cite{jin2022dataless} align models through parameter-space averaging under feature constraints. Recent studies have extended these paradigms to multi-task learning~\cite{caruana1997multitask,vandenhende2021multi,zhang2022survey}. LoRA merging involves combining multiple LoRA~\cite{hu2022lora} adapters to create a more versatile model. For example, Meta LoRA~\cite{li2025meta} extends the idea of low-rank adaptation by incorporating meta-learning methods, enabling models to quickly adapt to new tasks with minimal data and computation. TIES-MERGING~\cite{yadav2023ties} efficiently merges models by addressing interference in parameter values, involving pruning of task vectors, selection of parameter symbols, and disjoint fusion to improve multitask performance. This was further enhanced by DARE-TIES~\cite{yu2024language}, which applies sparsification and rescaling to task vectors before TIES merging for improved stability. KnOTS~\cite{stoica2024knots} introduces SVD-based parameter-space alignment to enable seamless fusion of conflicting LoRA adapters without retraining. Based on these methods, our approach employs task vectors as differentiable guidance signals to establish dynamic equilibrium among competing optimization objectives across tasks.

\subsection{In-Context Learning}
In-context learning has emerged as a novel paradigm in machine learning~\cite{rubin2021learning,dong2022survey,wies2023learnability, yan20243dsceneeditor}. The foundational work by~\cite{brown2020language} first demonstrated the in-context learning capability of large language models (LLMs) when provided with a limited number of examples. Building upon this, MetaICL~\cite{min2021metaicl} integrates tasks into the ICL format, enabling models to achieve performance comparable to direct fine-tuning without extensive parameter adjustments. LaMDA~\cite{thoppilan2022lamda} emphasizes the command tuning of the model for a better understanding of task descriptions. To address ICL's limitations in multi-step reasoning,~\cite{wei2022chain} introduced Chain-of-Thought (CoT) prompting, which decomposes tasks into intermediate reasoning steps. Building on this,~\cite{press2022measuring} proposed Self-Ask, a hierarchical prompting framework that breaks complex queries into sub-questions. The fusion of ICL with LoRA emerged as a breakthrough for balancing adaptability and efficiency. In scenario understanding, IC-LoRA~\cite{huang2024context} is a novel approach that leverages in-context learning to adapt models to various tasks without extensive fine-tuning. At the same time, ICM-LoRA~\cite{shao2025context} extends the concept of IC-LoRA by incorporating meta-learning and conditional variational autoencoders (CVAE). In this paper, our framework acquires task-specific adaptation vectors through in-context learning by modeling the relationship between LoRA weight perturbations and context demonstrations.

\subsection{Meta Learning}
Meta learning, or ``learning to learn''~\cite{brazdil2008metalearning}, has garnered significant attention in the machine learning community~\cite{wang2024robot,vanschoren2018meta,finn2017model,li2017meta,lee2019meta,hsu2018unsupervised}. Early breakthroughs, such as model-agnostic meta-learning (MAML)~\cite{finn2017model}, demonstrated that initializing model parameters in a task-agnostic basin enables rapid adaptation with a few gradient steps. In the context of few-shot learning, Matching Networks~\cite{vinyals2016matching} utilizes a siamese network architecture to match new examples with previously seen ones. At the same time, Relationnet~\cite{sung2018learning} proposes a two-branch relational network that performs a few-shot learning by learning to compare query images with a small number of labeled sample images. Recent advances have focused on improving the efficiency and scalability of meta-learning \cite{remondino2023critical, yan2025learning}. Developing first-order meta learning algorithms, such as FOMAML and Reptile~\cite{nichol2018first}, has reduced the computational overhead associated with second-order derivatives in MAML. To develop robust LLMs adaptable to unseen tasks, meta-training approaches like MetaICL~\cite{min2021metaicl} and MetaICT~\cite{chen2021meta} have been proposed, involving meta-training pre-trained LLMs models on diverse tasks through in-context multi-task fine-tuning and evaluating on disjoint test sets. Building on these, MAML-en-LLM~\cite{sinha2024maml} can learn truly generalizable parameters that perform well in disjoint tasks and adapt to unseen tasks. Our work aims to merge meta-learning principles with in-context learning in the context of LoRA adapters.

\begin{figure*}
    \centering
    \includegraphics[width=1\linewidth]{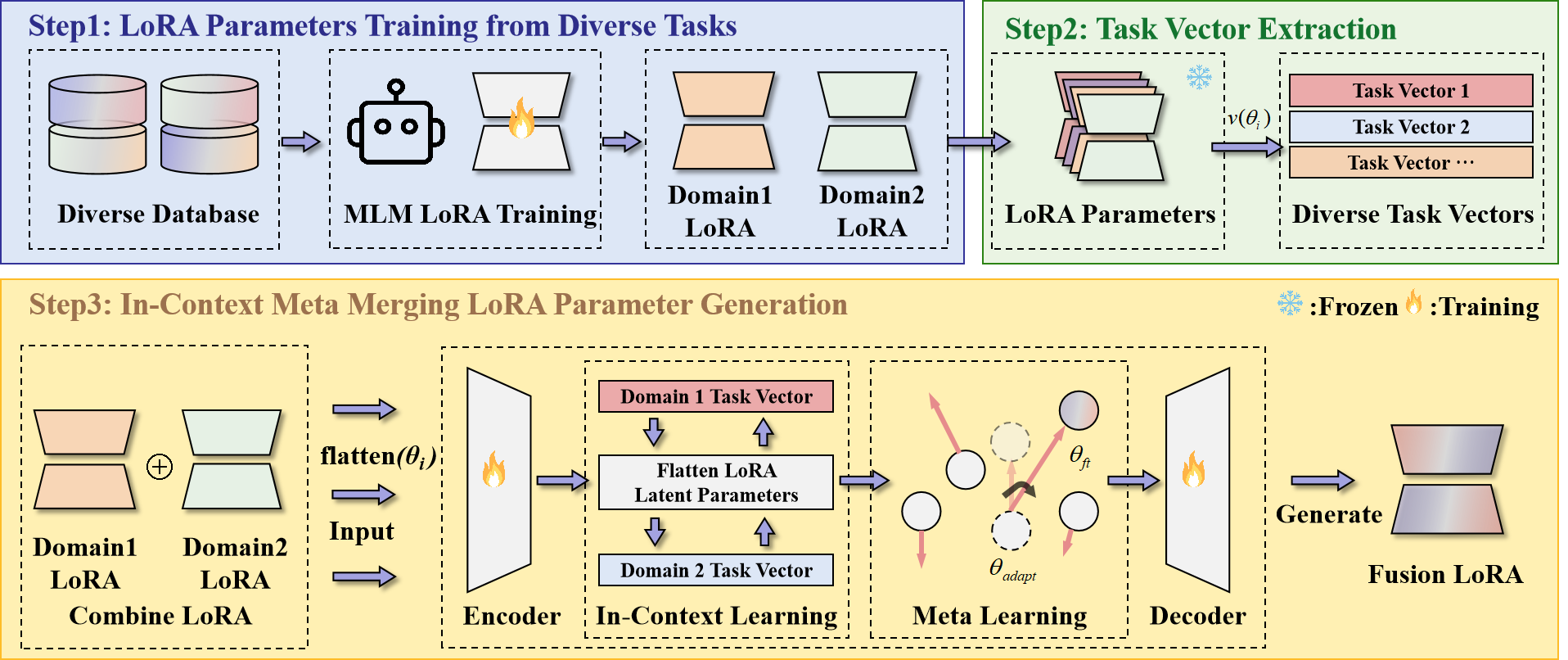}
    \caption{\textbf{Overview of the In-Context Meta Fusion LoRA framework.}
    \textbf{Step 1:} LoRA adapters are trained on a diverse set of tasks using masked language modeling (MLM), resulting in task-specific LoRA weights.
    \textbf{Step 2:} For each task, a task vector is obtained by computing the parameter difference between the fine-tuned and pre-trained LoRA adapters, providing a compact representation of task-specific knowledge.
    \textbf{Step 3:} The Fusion VAE leverages in-context learning and meta-learning, taking multiple task vectors and LoRA parameters as input to model latent task relationships. For each parameter dimension, the input $\theta_{\mathrm{adapt}}$ refers to the value of that specific dimensions relevant to the task in the flattened LoRA parameters; the VAE then adjusts $\theta_{\mathrm{adapt}}$ to generat e the task-specialized parameter $\theta_{\mathrm{ft}}$ suitable for the target task. The encoder-decoder structure enables efficient parameter fusion and adaptation.}
    \label{fig:framework-label}
\end{figure*}

\section{Methodology}

\subsection{Problem Settings}
We focus on \textbf{In-Context Meta LoRA Fusion}, aiming to address the multitask fusion problem for LoRA fine-tuned models by fusing the latent spaces of task-specific Variational Autoencoders (VAEs). Consider \(n\) isomorphic models \(\{f^{(1)}, f^{(2)}, \dots, f^{(n)}\}\), each fine-tuned on a distinct task \(\mathcal{T}_i\) (e.g., image classification, object detection) from a shared pre-trained checkpoint \(f^{\text{pt}}\). The parameters of the pre-trained model are denoted as \(\theta^{\text{pt}} = \{W_j^{\text{pt}}\}_{j=1}^L\), where \(W_j^{\text{pt}}\) represents the weights of the \(j\)-th layer and \(L\) is the total number of layers. Each task-specific LoRA model is associated with a VAE (\(\text{VAE}_i\)), which encodes task-specific LoRA parameters into a latent space, enabling cross-task fusion in this latent space. This approach aims to improve the recognition accuracy of the merged model by effectively integrating task-specific knowledge.

\subsection{Task Vector Extraction}
\label{sec:task_vector_extraction}
Following \cite{hendel2023context,li2025when}, we define a \textbf{task vector} $\Delta\mathbf{v}_{\mathcal{T}_i}$ for task $\mathcal{T}_i$ as the element-wise difference between the final-layer output tokens of the fine-tuned and pre-trained models:
\begin{equation}
    \Delta\mathbf{v}_{\mathcal{T}_i} = \mathbf{z}_{\mathcal{T}_i}^{*} - \mathbf{z}^{(0)},
     \label{eq:task_vector}
\end{equation}
where $\mathbf{z}^{(0)}$ denotes the output tokens from the last layer of the pre-trained model, and $\mathbf{z}_{\mathcal{T}_i}^{*}$ denotes the output tokens from the last layer after fine-tuning on task $\mathcal{T}_i$. The task vector $\Delta\mathbf{v}_{\mathcal{T}_i}$ thus captures the knowledge and adaptation required for the model to specialize in task $\mathcal{T}_i$ based on its final representations.For notational brevity, we use $\mathbf{v}_{\mathcal{T}i}$ to denote the task vector $\Delta\mathbf{v}{\mathcal{T}_i}$ as defined in Eq.~\ref{eq:task_vector}.

\subsection{Fusion VAE for Latent Space Encoding}
We introduce a \textbf{Fusion VAE}  method for encoding LoRA parameters in the latent space, enabling effective cross-task fusion. This approach differs from traditional CVAE methods in its handling of task semantics. Specifically, it represents each task's LoRA parameters \(\mathbf{l}^{(i)} = \text{flatten}(\tau^{(i)}) \in \mathbb{R}^D\) in the latent space to facilitate subsequent fusion operations.

Let $\phi$ denote the parameters of the encoder (inference network), and $\theta$ denote the parameters of the decoder (generative network) in the VAE framework. Throughout this paper, we use $\theta_i$ to represent the LoRA parameters associated with task $\mathcal{T}_i$, and $\psi$ to denote the decoder network parameters when necessary.

We employ a variational encoder to model the distribution over the latent variable $\mathbf{z}$ given both the LoRA parameters and the corresponding task vector. Specifically, we introduce an approximate posterior distribution $q_\phi(\mathbf{z} \mid \mathbf{l}^{(i)}, \mathbf{v}_{\mathcal{T}_i})$. This distribution is learned to approximate the true posterior $p(\mathbf{z} \mid \mathbf{l}^{(i)}, \mathbf{v}_{\mathcal{T}_i})$ over the latent variable $\mathbf{z}$ conditioned on the observed inputs.

The encoder of the Fusion VAE takes the concatenation of $\mathbf{l}^{(i)}$ and $\mathbf{v}_{\mathcal{T}_i}$ as input, and outputs the mean $\mu_\phi$ and variance $\sigma_\phi^2$ of a Gaussian distribution in the latent space:
\begin{align}
\mathbf{z} \sim q_\phi(\mathbf{z} \mid \mathbf{l}^{(i)},& \mathbf{v}_{\mathcal{T}_i}) = 
\notag
\\&
\mathcal{N}\left(\mu_\phi([\mathbf{l}^{(i)}; \mathbf{v}_{\mathcal{T}_i}]),\, \sigma_\phi^2([\mathbf{l}^{(i)}; \mathbf{v}_{\mathcal{T}_i}]) \mathbf{I}\right),
\end{align}
where $[\mathbf{l}^{(i)}; \mathbf{v}_{\mathcal{T}_i}]$ denotes the concatenation of the two input vectors, and $\mathbf{I}$ is the identity matrix that ensures the covariance is diagonal.

The decoder reconstructs the LoRA parameters from a latent vector \(\mathbf{z} \in \mathbb{R}^k\) and \(\mathbf{v}_{\mathcal{T}_i}\):
\begin{align}
\hat{\mathbf{l}}^{(i)} \sim p_\theta(\mathbf{l}^{(i)} \mid \mathbf{z}, &\mathbf{v}_{\mathcal{T}_i}) = 
\notag
\\&\mathcal{N}\left(\hat{\mu}_\theta([\mathbf{z}; \mathbf{v}_{\mathcal{T}_i}]), \hat{\sigma}_\theta^2([\mathbf{z}; \mathbf{v}_{\mathcal{T}_i}]) \mathbf{I}\right).
\end{align}
where $\hat{\mu}$ and $\hat{\sigma}$ denote the mean and variance outputs of the decoder network parameterized by $\theta$, given the concatenated latent vector and task vector as input.

To balance reconstruction accuracy and latent space regularization, we maximize the Evidence Lower Bound (ELBO):
\begin{align}
\mathcal{L}_i = \mathbb{E}_{q_\phi} &\left[ \log p_\theta(\mathbf{l}^{(i)} \mid \mathbf{z}, \mathbf{v}_{\mathcal{T}_i}) \right] 
\notag
\\&- \text{KL}\left( q_\phi(\mathbf{z} \mid \mathbf{l}^{(i)}, \mathbf{v}_{\mathcal{T}_i}) \parallel \mathcal{N}(0, \mathbf{I}) \right).
\label{eq:elbo}
\end{align}
In the KL divergence term, $\mathcal{N}(0, \mathbf{I})$ denotes the standard normal prior distribution (i.e., mean zero and identity covariance), which regularizes the latent space towards a unit Gaussian.

\subsection{In-Context Meta Learning}
Traditional meta-learning often requires explicit task boundaries and repeated retraining, which is impractical for large models and dynamic task settings. In-context meta learning overcomes these limitations by leveraging contextual information (e.g., LoRA parameters and task vectors) at inference time, enabling rapid adaptation and flexible fusion without costly retraining. This approach improves scalability and generalization to new or evolving tasks by utilizing latent representations learned from previous experiences.

Our methodology focuses on an advanced meta-learning framework designed to optimize the integration of multiple task-specific VAEs. This framework consists of two main components: task-specific adaptation and meta-parameter updating.

During the task-specific adaptation phase, each task-specific VAE is adapted to its corresponding task. The encoder \(E_{\phi}\) processes the task-specific LoRA parameters \(\theta_i\) and generates a latent distribution, which is defined by its mean \(\mu_i\) and logarithmic variance \(\log\sigma_i^2\):
\begin{equation}
(\mu_i, \log\sigma_i^2) = E_{\phi}(\theta_i).
\end{equation}
From this latent distribution, a sample vector $\mathbf{z}_i \in \mathbb{R}^d$ is drawn, where $d$ denotes the dimensionality of the latent space. This sample is then used by the decoder $D_{\psi}$ to generate an initial parameter estimate $\theta_i^{\text{init}}$:
\begin{equation}
\theta_i^{\text{init}} = D_{\psi}(\mathbf{z}_i).
\end{equation}
This initial estimate is refined through $K$ steps of gradient-based adaptation to minimize the task-specific reconstruction loss:
\begin{equation}
\theta_i^{(0)} = \theta_i^{\text{init}}
\end{equation}
\begin{equation}
\theta_i^{(k)} = \theta_i^{(k-1)} - \beta \nabla_{\theta} \mathcal{L}_{\text{task}}(\theta_i^{(k-1)}, \mathcal{D}_i), 
\label{eq:adapt_theta}
\end{equation}
where $k$ =  $1,\dots, K$.

The final adapted parameter is denoted as:
\begin{equation}
\theta_i^{\text{adapt}} = \theta_i^{(K)}
\end{equation}

In the meta-parameter updating component, the focus is on enhancing the fusion capabilities of the VAE parameters. The reconstruction loss is calculated to evaluate how accurately the original parameters can be reconstructed from the adapted latent representation:
\begin{equation}
\mathcal{L}_{\text{recon}} = \text{MSE}(\theta_i, \theta_i^{\text{recon}}).
\end{equation}
This loss measures the discrepancy between the original and reconstructed parameters. Additionally, the KL divergence loss is computed to regularize the latent distribution, ensuring it aligns with a standard normal prior:
\begin{align}
\mathcal{L}_{\text{KL}}&=
\notag
\\&
-\frac{1}{2} \sum \left( 1 +\log\sigma_i^{\text{adapt}^2} - (\mu_i^{\text{adapt}})^2 - \sigma_i^{\text{adapt}^2} \right).
\label{eq:KL_loss}
\end{align}
This loss encourages the latent space to maintain a structure that is consistent with a normal distribution, which is beneficial for generating new data.

These two losses are combined into a meta-loss, which guides the optimization of the VAE parameters:
\begin{equation}
\mathcal{L}_{\text{meta}} = \mathcal{L}_{\text{recon}} + \lambda_{\text{KL}} \cdot \mathcal{L}_{\text{KL}},
\label{eq:metaloss}
\end{equation}
where $\lambda_{\text{KL}} > 0$ is a weighting coefficient that balances the regularization term (KL divergence) and the reconstruction objective. This formulation ensures that the learned latent space is both informative for reconstruction and regularized to match the prior distribution, which is essential for robust task fusion.

In particular, the meta-loss in Eq.~\ref{eq:metaloss} shares a close connection with the evidence lower bound (ELBO) in Eq.~\ref{eq:elbo}, which forms the theoretical foundation of variational autoencoders. While Eq.~\ref{eq:elbo} describes the objective for reconstructing the original LoRA parameters for a single task, Eq.~\ref{eq:metaloss} extends this principle to the meta-learning setting, where the optimization is performed over multiple tasks to promote generalization and effective fusion across the task batch.

The encoder and decoder parameters are updated based on this meta-loss:
\begin{equation}
\phi \leftarrow \phi - \gamma \nabla_{\phi} \mathcal{L}_{\text{meta}},
\end{equation}
\begin{equation}
\psi \leftarrow \psi - \gamma \nabla_{\psi} \mathcal{L}_{\text{meta}}.
\end{equation}
Here, $\gamma$ denotes the meta step size used for updating the VAE parameters $(\phi, \psi)$ across tasks, in contrast to the inner-loop adaptation step size $\beta$ in Eq.~\ref{eq:adapt_theta}, which controls the gradient-based update for task-specific parameter adaptation. This separation ensures that meta-learning proceeds at an appropriate scale, independent of the rapid adaptation dynamics within each task.

This meta-learning process enables the Fusion VAE to effectively assimilate knowledge from multiple tasks, thereby significantly enhancing the performance of the merged model across various recognition tasks. By iteratively adapting to individual tasks and updating the model based on a meta-loss, the VAE develops a robust and generalizable representation of the latent space. This representation captures the essential features of each task while maintaining the model's ability to merge them effectively.

\begin{algorithm}[tb]
\caption{In-Context Meta LoRA Fusion Meta-Learning Procedure}
\label{alg:meta-lora-merging}
\begin{algorithmic}[1]
\REQUIRE Task set $\mathcal{T}$, initial VAE parameters $\phi, \psi$, meta step size $\gamma$, adaptation step size $\beta$, KL weight $\lambda_{\text{KL}}$
\FOR{$\mathrm{metaIter} = 1$ to $M$}
    \STATE Sample a batch of tasks $\{\mathcal{T}_i\}$ from $\mathcal{T}$
    \FOR{each task $\mathcal{T}_i$ in the batch}
        \STATE $\theta_i \gets \mathrm{ObtainLoRAParams}(\mathcal{T}_i)$
        \STATE $v_{\mathcal{T}_i} \gets \mathrm{ComputeTaskVector}(\mathcal{T}_i)$ \quad /* Eq.~\eqref{eq:task_vector} */
        \STATE $(\mu_i, \log\sigma_i^2) \gets E_\phi([\theta_i; v_{\mathcal{T}_i}])$
        \STATE $z_i \sim \mathcal{N}\bigl(\mu_i, \mathrm{diag}(\sigma_i^2)\bigr)$
        \STATE $\theta_i^{\mathrm{init}} \gets D_\psi([\!z_i; v_{\mathcal{T}_i}])$
        \STATE $\theta_i^{\mathrm{adapt}} \gets \theta_i^{\mathrm{init}} \;-\; \beta \,\nabla_{\theta_i^{\mathrm{init}}}\,\mathcal{L}_{\text{task}}\bigl(\theta_i^{\mathrm{init}}, \mathcal{D}_i\bigr)$
        \STATE $(\mu_i', \log\sigma_i'^2) \gets E_\phi([\theta_i^{\mathrm{adapt}}; v_{\mathcal{T}_i}])$
        \STATE $\theta_i^{\mathrm{recon}} \gets D_\psi([\!z_i; v_{\mathcal{T}_i}])$
        \STATE $\mathcal{L}_{\mathrm{recon}} \gets \mathrm{MSE}(\theta_i, \theta_i^{\mathrm{recon}})$
        \STATE $\mathcal{L}_{\mathrm{KL}} \gets \mathrm{KL}\!\Bigl(\mathcal{N}(\mu_i', \sigma_i'^2),\, \mathcal{N}(0, I)\Bigr)$ /* Eq.~\eqref{eq:KL_loss} */
        \STATE $\mathcal{L}_{\mathrm{meta}} \gets \mathcal{L}_{\mathrm{recon}} + \lambda_{\text{KL}} \,\mathcal{L}_{\mathrm{KL}}$
    \ENDFOR
    \STATE $\phi \;\gets\; \phi \;-\; \gamma \,\nabla_{\phi}\!\Bigl(\sum_i \mathcal{L}_{\mathrm{meta}}\Bigr)$
    \STATE $\psi \;\gets\; \psi \;-\; \gamma \,\nabla_{\psi}\!\Bigl(\sum_i \mathcal{L}_{\mathrm{meta}}\Bigr)$
\ENDFOR
\end{algorithmic}
\end{algorithm}

\section{Experiments}
In this section, we discuss the experimental setup and implementation details. Additionally, we report the performance of different models on vision and language tasks.

\subsection{Experiment Settings}
\label{sec:setup}
\textbf{Baselines.} We adopted the original model with LoRA~\cite{hu2022lora}, which is generated by Model Soup~\cite{wortsman2022model}, RegMean~\cite{jin2022dataless}, TA~\cite{ilharco2022editing}, KnOTS~\cite{stoica2024knots}, TIES~\cite{yadav2023ties}, and DARE~\cite{yu2024language} as baselines to compare with our method, aiming to assess its advantages on various tasks.

\noindent \textbf{Datasets.} For the computer vision task, we focus on the target detection task and the multimodal question-answering task. So we employ the VOC 2012 dataset~\cite{everingham2010pascal} and the ScienceQA dataset~\cite{lu2022learn}.
For the language task, we utilize The Pile~\cite{gao2020pile} as the language modeling task dataset.

\noindent \textbf{Data Preparation.} During model fine-tuning, a sequence of LoRA matrices $\{L_t\}_{t=1}^T$ with varying ranks $r$ is generated for multiple diverse tasks, where $T$ denotes the number of fine-tuning steps. Each matrix $L_t \in \mathbb{R}^{m \times n}$ is flattened into a one-dimensional array $l_t \in \mathbb{R}^{m \times n}$ to enable alignment with the task vector $v_{\text{task}}$. These flattened LoRA parameters, combined with their corresponding task vectors, form the training data set $\{(v_{\text{task}}, l_t)\}$ for the \textbf{self-designed VAE}. This data set captures the relationship between task-specific representations and LoRA parameter distributions. For the two selected tasks, we obtain their respective LoRA parameter sequences $\{L_{t1}\}_{t=1}^{T1}$ and $\{L_{t2}\}_{t=1}^{T2}$, flatten them into arrays $l_{t1}$ and $l_{t2}$, and construct data sets $D_1 = \{(v_{\text{task1}}, l_{t1})\}$ and $D_2 = \{(v_{\text{task2}}, l_{t2})\}$. These data sets are merged into a combined data set $D_{\text{combined}} = D_1 \cup D_2$. During VAE training, batches are randomly drawn from $D_{\text{combined}}$, enabling the VAE to learn from the different LoRA parameters of both tasks and forming the basis for subsequent parameter generation and model customization.


\begin{figure}
    \centering
    \includegraphics[width=1\linewidth]{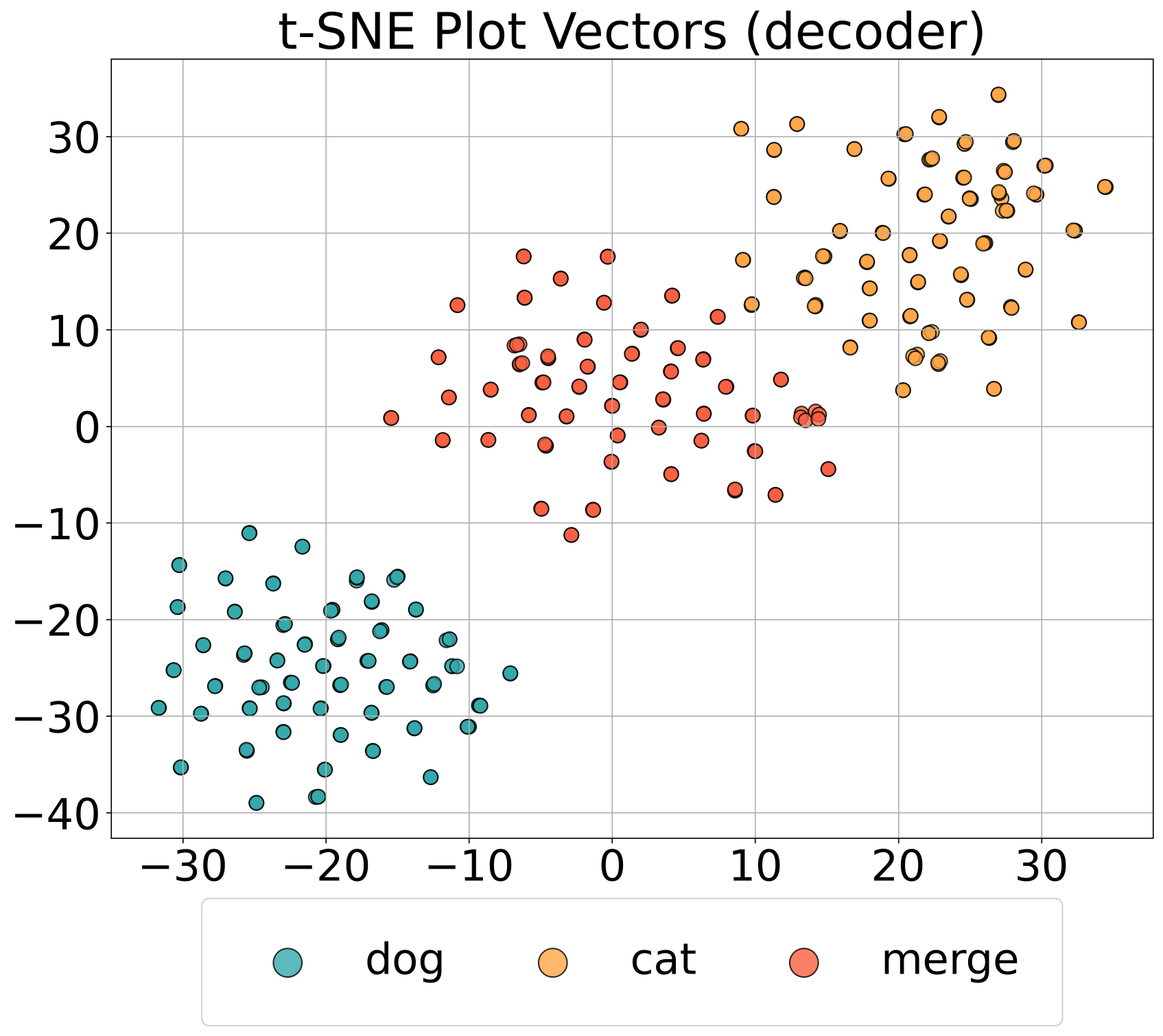}
    \caption{\textbf{t-SNE visualization of decoder task vectors for the dog, cat, and ICM-Fusion models.} The plot shows the fused vectors distributed between the dog and cat clusters, indicating that the merged model captures features from both individual tasks.}
    \label{fig:tsne-vae-catdog-decoder}
\end{figure}
\begin{table*}[ht] \small
\centering
\begin{tabular}{lccccccccc}
\toprule
~ & \multicolumn{8}{c}{\textbf{MAP50}}  \\
\multirow{-2}{*}{\textbf{Method}} 
    & Cat & Dog & Train & Bus & Bicycle & Horse & Sheep & Bird & Avg.  \\
\midrule
Original Model      & 0.94 & 0.92 & 0.89 & 0.87 & 0.85 & 0.83 & 0.81 & 0.76 & 0.86 \\
Original LoRA      & 0.95 & 0.92 & 0.91 & \textbf{0.92} & 0.88 & \textbf{0.90} & 0.85 & 0.82 & 0.89 \\\midrule
Model Soup~\cite{wortsman2022model}       & 0.93 & 0.92 & 0.88 & 0.84 & 0.85 & 0.85 & 0.82 & 0.75 & 0.84 \\ 
RegMean~\cite{jin2022dataless}          & 0.95 & 0.92 & 0.90 & 0.89 & 0.87 & 0.86 & 0.85 & 0.82 & 0.88 \\
TA  ~\cite{ilharco2022editing}                & 0.95 & 0.93 & 0.91 & 0.90 & 0.88 & 0.87 & 0.86 & 0.83 & 0.89 \\
SVD in Latent State~\cite{koren2008factorization} & 0.89 & 0.87 & 0.85 & 0.84 & 0.84 & 0.81 & 0.80 & 0.73 & 0.83 \\
KnOTS  ~\cite{stoica2024knots}             & 0.94 & 0.92 & 0.90 & 0.89 & 0.86 & 0.83 & 0.82 & 0.76 & 0.87 \\
TIES ~\cite{yadav2023ties}                & 0.93 & 0.91 & 0.88 & 0.87 & 0.85 & 0.83 & 0.81 & 0.77 & 0.86 \\
DARE-TIES  ~\cite{yu2024language}         & 0.92 & 0.90 & 0.87 & 0.86 & 0.84 & 0.82 & 0.80 & 0.76 & 0.85 \\
\textbf{ICM-Fusion (Ours)} & \textbf{0.96} & \textbf{0.93} & \textbf{0.92} & 0.91 & \textbf{0.90} &0.88 & \textbf{0.87} & \textbf{0.85} & \textbf{0.90} \\
\bottomrule
\end{tabular}
\caption{\textbf{Results on object detection with rank $r=8$.} Compared to the original LoRA, ICM-Fusion achieves significant improvement in most samples and demonstrates enhanced overall detection performance.}
\label{tab:merge_results1}
\end{table*}

\begin{table}[htbp]
\resizebox{1\linewidth}{!}{
\begin{tabular}{lcccccccc}
\toprule
\textbf{Method} & NAT   & SOC   & LAN   & TXT   & IMG   & NO    & \multicolumn{2}{c}{Avg.} \\
\midrule
Original Model     & 89.32 & 96.08 & 85.32 & 88.95 & 87.17 & 88.43 & 89.38 \\
Original LoRA     & \textbf{89.34} & \textbf{96.10} & 85.30 & 88.94 & 87.16 & 88.44 & 89.38 \\\midrule
Model Soup~\cite{wortsman2022model}         & 87.61 & 93.02 & 85.33 & 82.91 & 86.83 & 85.23 & 86.82 \\
RegMean~\cite{jin2022dataless}            & 88.91 & 95.82 & 85.11 & 88.42 & 87.08 & 88.01 & 87.73 \\
TA~\cite{ilharco2022editing}                 & 89.01 & 95.96 & 85.21 & 88.70 & 87.12 & 88.15 & 87.86 \\
SVD~\cite{koren2008factorization}                & 88.85 & 95.68 & 85.18 & 88.31 & 87.03 & 88.05 & 87.85 \\
KnOTS~\cite{stoica2024knots}              & 89.10 & 96.03 & 85.27 & 88.81 & 87.16 & 88.34 & 88.12 \\
TIES~\cite{yadav2023ties}               & 88.98 & 95.94 & 85.24 & 88.65 & 87.14 & 88.22 & 88.03 \\
DARE-TIES~\cite{yu2024language}          & 89.12 & 96.05 & 85.29 & 88.83 & 87.16 & 88.38 & 88.14 \\
\textbf{ICM-Fusion (Ours)}   & 89.33 & 96.08 & 85.32 & \textbf{88.96} & \textbf{87.17} & \textbf{88.45} & \textbf{89.39} \\ 

\bottomrule
\end{tabular}}
\vspace{-0.3cm}
\caption{\textbf{Result on ScienceQA.} Question classes: NAT = natural science, SOC = social science, LAN = language science, TXT = text context, IMG = image context, NO = no context.}
\label{tab:SQA}
\vspace{-0.4cm}
\end{table}
\subsection{Main Results}
\label{sec:main-results}We conducted experiments on image object detection on Florence2~\cite{xiao2024florence}, visual question answering on LLAVA-v1.5~\cite{liu2024improved}, and language modeling tasks on Llama3~\cite{grattafiori2024llama} to demonstrate the reliability and generalizability of our method.

\noindent \textbf{Results on Vision Tasks.}
As shown in Table~\ref{tab:merge_results1}, we compare various model fusion methods on object detection tasks with multiple categories. Our ICM-Fusion achieves the highest MAP50 scores and shows improvement in the relevant metrics. Although detection performance decreases in some samples, the overall results remain superior to existing methods.
This indicates that ICM-Fusion can enhance the capability of the fused model and optimize its performance when task vectors point in similar directions. Although affected by opposing vectors, ICM-Fusion effectively mitigates their negative impact. As shown in Figure~\ref{fig:tsne-vae-catdog-decoder}, the t-SNE visualization~\cite{maaten2008visualizing} shows that the sampled points of the fused model lie between the two domains, enabling it to adapt to both domains simultaneously.

As shown in Table~\ref{tab:SQA}, For the visual question answering (VQA) task, while fine-tuned model scores may vary due to non-robust benchmark evaluations, ICM-Fusion achieves a consistent and significant improvement in the average score across all tasks post-fusion compared to existing methods. This result demonstrates its capability to enhance the reasoning performance of merged models. Moreover, since LLaVA-v1.5~\cite{liu2024improved} fine-tunes only the decoder while Florence-2~\cite{xiao2024florence} requires fine-tuning both the encoder and decoder, there are certain structural differences between the fused models. Both tasks demonstrate that ICM-Fusion can robustly improve the performance of the merged model, indicating that ICM-Fusion is adaptable to LoRA-based fusion of models with different architectures.

\noindent \textbf{Few-shot Results.}
\noindent Table~\ref{tab:longtail_merge_pipelines} reports the results in few-shot settings. We evaluated MAP50 for several long-tail categories, Sofa, Airplane, Motorbike, and Dining Table, under two data regimes: 0\% (no additional training data) and +10\% (with a small amount of additional data). Across all methods, performance is limited when no extra data is provided. However, when 10\% more data is available, all methods exhibit improvements, with ICM-Fusion demonstrating the most significant gains and achieving the highest MAP50 scores in the few-shot category. This demonstrates that ICM-Fusion can still generalize to the target domain even in few-shot learning scenarios.
\begin{table*}[htbp] \small
\centering
\begin{tabular}{lcccccccc}
\toprule
 & \multicolumn{2}{c}{Sofa} & \multicolumn{2}{c}{Aeroplane} & \multicolumn{2}{c}{Motorbike} & \multicolumn{2}{c}{Dining Table} \\
\multirow{-2}{*}{\textbf{Method}}   & 0\% & +10\% & 0\% & +10\% & 0\% &      +10\% & 0\% & +10\% \\
\midrule
Original Model      & 0.00 & 0.79 & 0.00 & 0.75 & 0.00 & 0.73 & 0.00 & 0.76 \\
Original LoRA(Cat)  & 0.00 & 0.83 & 0.00 & 0.77 & 0.00 & 0.75 & 0.00 & 0.81 \\\midrule
Model Soup~\cite{wortsman2022model}          & 0.00 & 0.78 & 0.00 & 0.74 & 0.00 & 0.72 & 0.00 & 0.78 \\
RegMean~\cite{jin2022dataless}             & 0.00 & 0.81 & 0.00 & 0.76 & 0.00 & 0.74 & 0.00 & 0.80 \\
TA~\cite{ilharco2022editing}                  & 0.00 & 0.82 & 0.00 & 0.77 & 0.00 & 0.75 & 0.00 & 0.81 \\
SVD in Latent State ~\cite{koren2008factorization}& 0.00 & 0.77 & 0.00 & 0.73 & 0.00 & 0.71 & 0.00 & 0.78 \\
KnOTS~\cite{stoica2024knots}               & 0.00 & 0.80 & 0.00 & 0.75 & 0.00 & 0.73 & 0.00 & 0.80 \\
TIES~\cite{yadav2023ties}                & 0.00 & 0.79 & 0.00 & 0.74 & 0.00 & 0.72 & 0.00 & 0.79 \\
DARE-TIES~\cite{yu2024language}           & 0.00 & 0.78 & 0.00 & 0.74 & 0.00 & 0.72 & 0.00 & 0.78 \\
\textbf{CM-Fusion (Ours)}I& 0.00 & 0.85 & 0.00 & 0.80 & 0.00 & 0.79 & 0.00 & 0.85 \\
\bottomrule
\end{tabular}
\caption{\textbf{MAP50 scores for long-tail classes.} (Sofa, Aeroplane, Motorbike, Dining Table) for different model fusion pipelines under two data percentages.}
\label{tab:longtail_merge_pipelines}
\end{table*}
\begin{table*}[ht]\small
\centering
\begin{tabular}{l
                cc
                cc
                cc
                cc
                cc
                cc}
\toprule
~ & \multicolumn{2}{c}{ArXiv}
  & \multicolumn{2}{c}{Books}
  & \multicolumn{2}{c}{Ubuntu}
  & \multicolumn{2}{c}{Wikipedia}
  & \multicolumn{2}{c}{Gutenberg}
  & \multicolumn{2}{c}{Avg.} \\
\multirow{-2}{*}{\textbf{Method}}
    & PPL & BPC
    & PPL & BPC
    & PPL & BPC
    & PPL & BPC
    & PPL & BPC
    & PPL & BPC \\
\midrule
Original Model     & 7.76 & 0.44 & 7.67 & 0.51 & 10.00 & 0.59 & 6.07 & 0.46 & 8.75 & 0.63 & 8.45 & 0.53 \\
Original LoRA      & 6.70 & 0.41 & 7.10 & 0.48 & 9.67 & 0.59 & 5.58 & 0.44 & 8.60 & 0.62 & 7.53 & 0.51 \\\midrule
Model Soup~\cite{wortsman2022model}         & 7.00 & 0.43 & 7.08 & 0.48 & 9.67 & 0.59 & 5.56 & 0.45 & 8.61 & 0.63 & 7.58 & 0.52 \\
RegMean~\cite{jin2022dataless}            & 6.90 & 0.43 & 7.10 & 0.48 & 9.68 & 0.59 & 5.60 & 0.45 & 8.63 & 0.62 & 7.58 & 0.51 \\
TA~\cite{ilharco2022editing}                 & 6.85 & 0.43 & 7.09 & 0.48 & 9.67 & 0.59 & 5.57 & 0.45 & 8.62 & 0.62 & 7.56 & 0.51 \\
SVD in Latent State~\cite{koren2008factorization}& 6.84 & 0.43 & 7.13 & 0.48 & 9.69 & 0.59 & 5.60 & 0.45 & 8.66 & 0.63 & 7.58 & 0.52 \\
KnOTS~\cite{stoica2024knots}              & 6.78 & 0.43 & 7.13 & 0.48 & 9.70 & 0.58 & 5.61 & 0.45 & 8.68 & 0.62 & 7.58 & 0.51 \\
TIES~\cite{yadav2023ties}               & 6.79 & 0.43 & 7.14 & 0.48 & 9.71 & 0.59 & 5.62 & 0.45 & 8.69 & 0.63 & 7.59 & 0.52 \\
DARE-TIES~\cite{yu2024language}          & 6.81 & 0.43 & 7.16 & 0.48 & 9.72 & 0.58 & 5.63 & 0.45 & 8.70 & 0.62 & 7.60 & 0.51 \\
\textbf{ICM-Fusion (Ours)} & \textbf{6.73} & \textbf{0.42} & \textbf{7.07} & \textbf{0.47}
                & \textbf{9.64} & \textbf{0.58} & \textbf{5.55} & \textbf{0.43}
                & \textbf{8.57} & \textbf{0.61} & \textbf{7.51} & \textbf{0.50} \\
\bottomrule
\end{tabular}
\caption{Language modeling results on different datasets (ArXiv, Books, Ubuntu, Wikipedia, Gutenberg) for various model fusion methods. Metrics reported are Perplexity (PPL) and Bits-per-Character (BPC), and their average values.}
\label{tab:lm_eval}
\end{table*}

\noindent \textbf{Results on Language Tasks.}
To validate the generalization capability of ICM-Fusion across different modal tasks, we selected a pure language modeling task with LLAMA3~\cite{grattafiori2024llama} on the Pile dataset~\cite{gao2020pile}. 
To assess the effectiveness of ICM-Fusion in the context of language modeling, we conduct experiments across multiple established benchmarks. For consistency, we also conducted experiments on language modeling tasks using related methods, as shown in Table~\ref{tab:lm_eval}. For most samples, the PPL shows a decrease compared to the original LoRA before fusion, indicating that the fused LoRA effectively enhances the language modeling capability of the model. Although the BPC change before and after fine-tuning is relatively small, the fused model still achieves better BPC performance on most samples, demonstrating that ICM-Fusion adapts well to language modeling tasks.

Based on the results from Tables~\ref{tab:merge_results1},~\ref{tab:SQA}, and~\ref{tab:lm_eval}, ICM-Fusion demonstrates adaptability to fine-tune LoRA across different model architectures and multimodal tasks. This indicates that our method not only enhances the capabilities of pre-fusion domain-specific LoRA but also generalizes effectively to diverse models and cross-modal applications.

These results highlight the robustness and adaptability of ICM-Fusion in few-shot scenarios, particularly for rare classes where training samples are scarce. The substantial improvement over baseline fusion methods underscores its effectiveness for long-tail recognition under limited data conditions.

\subsection{Ablation Study}
To investigate the impact of source data hyperparameters on the performance of ICM-Fusion, we conducted ablation studies on the rank of LoRA (which determines the number of LoRA parameters) and the proportion of sampled data.

\noindent \textbf{Effect of LoRA Ranks.} Table~\ref{tab:lora_rank_ablation} presents the results of varying the LoRA rank $r$ in ICM-Fusion. We observe that MAP50 scores for representative categories (Cat, Dog, Bus) remain consistently high across different ranks. The results demonstrate that the performance of the fused model remains nearly constant across different LoRA ranks (parameter numbers). Thus, ICM-Fusion exhibits robustness to LoRA models with varying parameter numbers.
\begin{table}[ht] \small
\centering
\begin{tabular}{ll
                ccc
                ccc
                ccc}
\toprule
~ & ~ &  \multicolumn{3}{c}{ICM-Fusion} \\
\multirow{-2}{*}{\textbf{Rank}} & \multirow{-2}{*}{\textbf{Parameter}}
    & Cat & Dog & Bus
    \\
\midrule
$r=1$   & 241,241     & 0.96 & 0.93 & 0.91 \\
$r=2$   & 482,482     & 0.96 & 0.93 & 0.91 \\
$r=4$   & 964,964     & 0.96 & 0.93 & 0.91 \\
$r=8$   & 1,929,928   & 0.96 & 0.93 & 0.91 \\
$r=16$  & 3,859,856   & 0.96 & 0.93 & 0.91 \\
\bottomrule
\end{tabular}
\caption{\textbf{Impact of LoRA Rank and Parameter.} ICM-Fusion demonstrates robust performance across LoRA weights with different ranks and parameters.}
\label{tab:lora_rank_ablation}
\end{table}

\noindent \textbf{Effect of Data Sampling Rate. }As shown in Table~\ref{tab:long_tail_map50}, we further analyze the sensitivity of ICM-Fusion to the amount of available training data for several long-tail classes. With no additional data, MAP50 scores are near zero. However, as the proportion of training data increases from 10\% to 30\%, substantial and consistent improvements are observed across all long-tail classes. These results highlight the data efficiency of ICM-Fusion and its ability to benefit from even small increases in training samples rapidly. Therefore, even in scenarios with limited training samples, ICM-Fusion can still achieve domain enhancement.
\begin{table}[htbp]
\centering
\resizebox{\linewidth}{!}{
\begin{tabular}{lcccc}
\toprule
\textbf{Data Percent} & \textbf{Sofa} & \textbf{Aeroplane} & \textbf{Motorbike} & \textbf{Dining Table} \\
\midrule
0\%   & 0.00 & 0.00 & 0.00 & 0.00 \\
+10\% & 0.83 & 0.79 & 0.81 & 0.80 \\
+20\% & 0.85 & 0.82 & 0.83 & 0.83 \\
+30\% & 0.88 & 0.85 & 0.87 & 0.86 \\
\bottomrule
\end{tabular}
}
\caption{MAP50 scores for long-tail classes under different proportions of training data.}
\label{tab:long_tail_map50}
\end{table}

\label{sec:ablation-study}

\section{Conclusion}

We propose a unified framework for In-Context Meta LoRA Fusion, which systematically integrates task-specific LoRA parameters via a Fusion VAE and meta-learning. ICM-Fusion generates an optimal fused model architecture by performing meta-learning-guided task vector direction optimization in the latent space after VAE sampling. Our experiments demonstrate that the method not only reduces storage overhead but also adapts effectively across vision and language tasks, even when direct supervision or extensive training are not readily available. We further discuss potential applications, such as sharing knowledge across multiple domains and augmenting new tasks with minimal overhead, thereby enhancing robustness and versatility in multitask systems. We believe our theoretical and empirical investigation offers deeper insights into parameter-efficient model unification and lays promising groundwork for future studies in broader AI applications.

\section{Appendix}
\noindent \textbf{Training Strategies.} The VAE model uses a 12-layer 1D CNN architecture for both the encoder and decoder. Its loss function combines the Kullback-Leibler divergence (KLD) and reconstruction loss with the KLD weight set at 0.005, and further integrates a meta loss term to balance task-specific adaptation and latent space generalization dynamically. We conducted fine-tuning on a specific task using LoRA over a total of 100 epochs, retaining LoRA parameters from the final 50 epochs. The task vector was extracted from the last token of the last layer in CLIP~\cite{radford2021learning}. Subsequently, the VAE model underwent 4,000 epochs of training to ensure robust learning of the latent space. All experiments were performed on a single NVIDIA A800 GPU, with each experiment taking approximately 3 hours to complete.

\noindent \textbf{Evaluation Metrics.} We record different LoRA ranks (\( r \)) and parameter counts (\( P \)) across all experiments. For object detection tasks, we report Mean Average Precision (MAP) at intersection-over-union (IoU) thresholds of \( \text{IoU} = 0.5 \) and \( \text{IoU} = 0.75 \). Specifically, MAP@50 is defined as
\begin{equation}
    \label{eq:map50}
    \text{MAP@50} = \frac{1}{N} \sum_{i=1}^{N} \text{AP}_{i, 0.5},
\end{equation}
where $\text{AP}_{i, 0.5}$ denotes the average precision for the $i$-th class at an IoU threshold of 0.5, and $N$ represents the total number of classes. MAP@50 provides a single value summarizing the model's detection performance across all categories, considering a detection correct if the intersection over union between the predicted and ground-truth bounding boxes is at least 0.5. The intersection over union (IoU) between a predicted bounding box $B_p$ and a ground-truth bounding box $B_{gt}$ is calculated as
\begin{equation}
    \label{eq:iou}
    \text{IoU}(B_p, B_{gt}) = \frac{\text{Area}(B_p \cap B_{gt})}{\text{Area}(B_p \cup B_{gt})}.
\end{equation}

For language modeling tasks, we employ perplexity (\textbf{PPL}) and bits-per-character (\textbf{BPC}) as evaluation metrics, where lower values indicate better performance. Perplexity is defined as
\begin{equation}
    \label{eq:ppl}
    \text{PPL} = \exp\left(-\frac{1}{N} \sum_{i=1}^N \log P(x_i \mid x_{0: i-1})\right),
\end{equation}
where $x_i$ is the predicted token at inference step $i$ conditioned on previous tokens. Bits-per-character (BPC) is computed as
\begin{equation}
    \label{eq:bpc}
    \text{BPC} = \frac{\log_2(\text{PPL})}{|c_w|},
\end{equation}
where $c_w$ denotes the average number of characters per word in the generated sequence. These definitions ensure clarity and consistency in evaluating both object detection and language modeling performance.

\bibliography{aaai2026}

\end{document}